\UseRawInputEncoding
\documentclass[letterpaper]{article} 
\usepackage{aaai24}  
\usepackage{times}  
\usepackage{helvet}  
\usepackage{courier}  
\usepackage[hyphens]{url}  
\usepackage{graphicx} 
\usepackage{natbib}  
\usepackage{caption} 
\usepackage{algorithm}
\usepackage{algorithmic}

\usepackage{newfloat}
\usepackage{listings}

\usepackage{epsfig}
\usepackage{amsmath}
\usepackage{amssymb}
\usepackage{multirow}
\usepackage{diagbox}
\usepackage{marvosym}
\DeclareCaptionStyle{ruled}{labelfont=normalfont,labelsep=colon,strut=off} 
\floatstyle{ruled}
\newfloat{listing}{tb}{lst}{}
\floatname{listing}{Listing}
\title{Dynamic Weighted Combiner for Mixed-Modal Image Retrieval}
\author{Fuxiang Huang, Lei Zhang\textsuperscript{\Letter}, Xiaowei Fu, Suqi Song
}
\affiliations{
Learning Intelligence \& Vision Essential (LiVE) Group\\
School of Microelectronics and Communication Engineering, Chongqing University, China\\
{\tt \{huangfuxiang, leizhang, xwfu, songsuqi\}@cqu.edu.cn}
}

\usepackage{bibentry}
\begin{document}

\maketitle

\begin{abstract}
Mixed-Modal Image Retrieval (MMIR) as a flexible search paradigm has attracted wide attention. However, previous approaches always achieve limited performance, due to two critical factors are seriously overlooked. 1) The contribution of image and text modalities is different, but incorrectly treated equally. 2) There exist inherent labeling noises in describing users' intentions with text in web datasets from diverse real-world scenarios, giving rise to overfitting.
We propose a Dynamic Weighted Combiner (DWC) to tackle the above challenges, which includes three merits.
\underline{First}, we propose an Editable Modality De-equalizer (EMD) by taking into account the contribution disparity between modalities, containing two modality feature editors and an adaptive weighted combiner.
\underline{Second}, to alleviate labeling noises and data bias, we propose a dynamic soft-similarity label generator (SSG) to implicitly improve noisy supervision.
\underline{Finally}, to bridge modality gaps and facilitate similarity learning, we propose a CLIP-based mutual enhancement module alternately trained by a mixed-modality contrastive loss.  
Extensive experiments verify that our proposed model significantly outperforms state-of-the-art methods on real-world datasets. The source code is available at \url{https://github.com/fuxianghuang1/DWC}.
\end{abstract}

\section{1. Introduction}
\begin{figure}[t]
\centering
\includegraphics[width=0.42\textwidth]{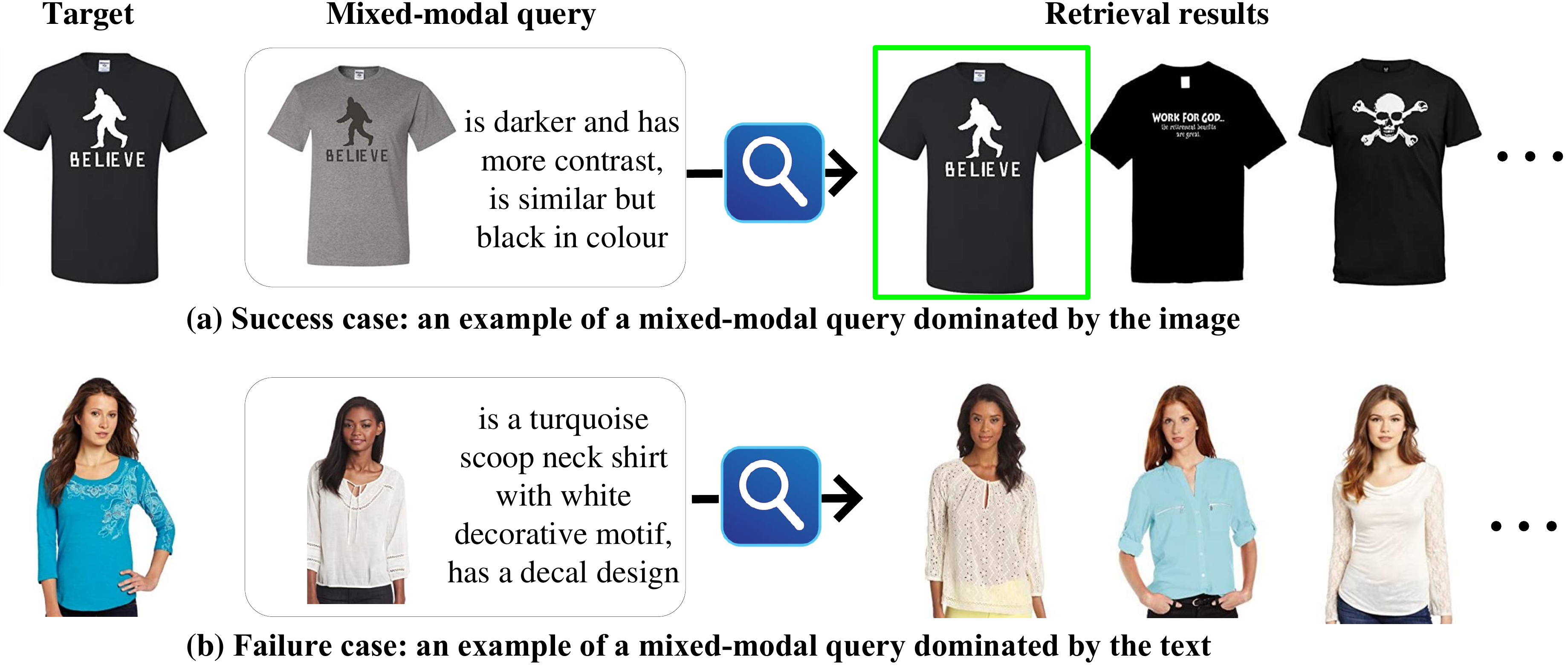}
\vspace{-1.2mm}
\caption{Overview of MMIR, i.e., image+text $\rightarrow$ image. The mixed-modal query is dominated by different modalities unequally. (a) Dominated by image modality. (b) Dominated by text modality.}
\label{fig1}
\vspace{-3.2mm}
\end{figure}

Image retrieval \cite{Chun2021, MultiSentence2021, Domain2021, Survey2022}, as a crucial computer vision task, aims to search for items of interest from the database. A key limitation of traditional image retrieval is the in-feasibility to precisely describe users' intentions (i.e., the concepts in users' minds) through a single image or a single text.
Therefore, to offer a flexible and intuitive user experience, a Mixed-Modal Image Retrieval (MMIR) paradigm as shown in Fig.~\ref{fig1} is explored, where the search intention is expressed with mixed modalities utilized to retrieve a target image.
Therefore, MMIR requires a synergistic understanding of both visual and linguistic content, which is an acknowledged challenge.

Most existing approaches \cite{vo2019composing, Hou2021ICCV, CLVC-Net2021, DATIR2021, FashionVLP2022} mainly focus on designing complex components to learn composite image-text representations. For example, \cite{vo2019composing} propose a TIRG model with a gated residual connection to modify partial image regions with text guidance. \cite{CLVC-Net2021} propose a CLVC-Net to combine local-wise and global-wise composition modules. 
\cite{FashionVLP2022} propose a vision-language transformer-based model, FashionVLP, to effectively incorporate multiple levels of fashion-related context. However, these methods usually achieve poor retrieval results, with almost no more than 20\% of queries retrieving the correct image in the top-1 rank.
\textit{We cannot help asking why it happened}?

\begin{figure}[t]
\centering
\begin{minipage}{4cm}
\includegraphics[scale=0.275]{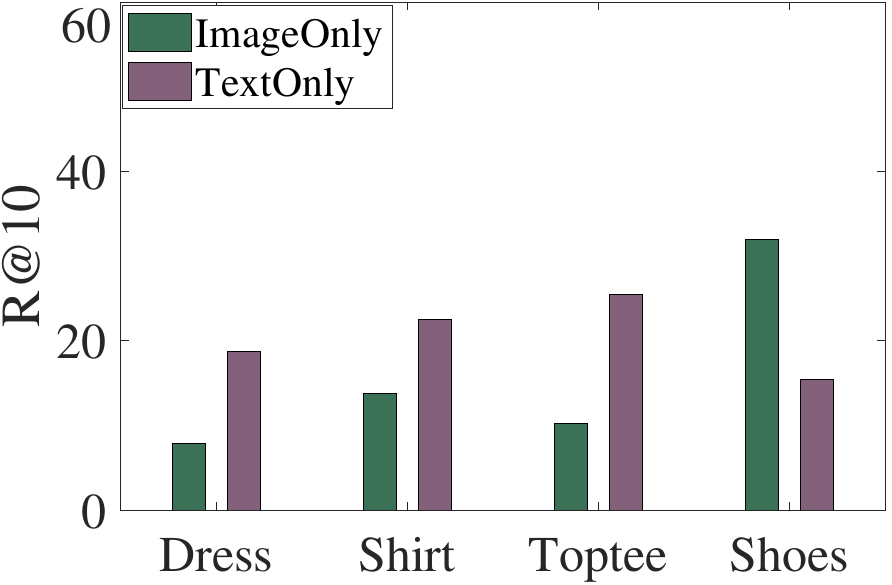}
\centerline{(a)}
\end{minipage}%
\hfill
\begin{minipage}{4cm}
\includegraphics[scale=0.275]{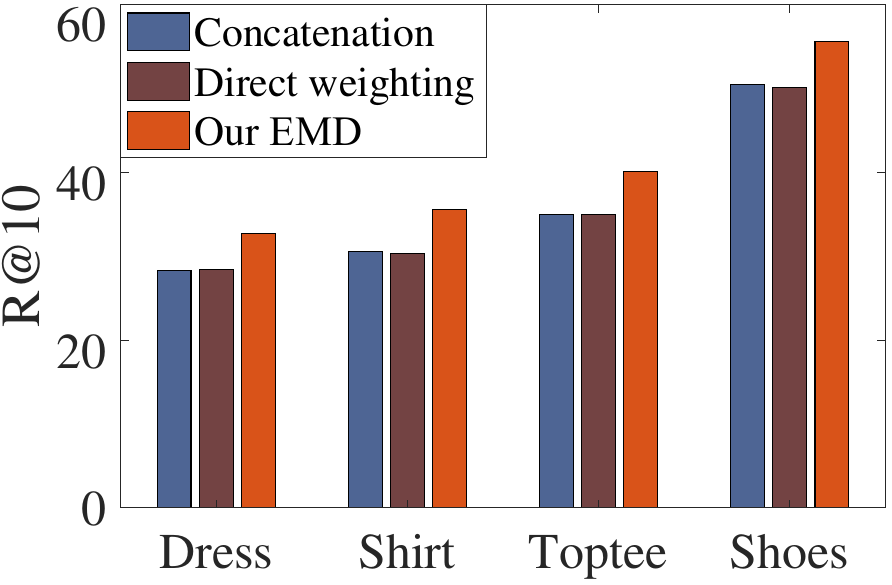}
\centerline{(b)}
\end{minipage}%
\vspace{-1.2mm}
\caption{ (a) Single-modal query.  (b) Mixed-modal query with different composition methods. \textit{Concatenation} means concatenating the image and text modality features.
\textit{Direct weighting} refers to the naive modality weighted method in \cite{PL2022}.}
\label{fig2}
\vspace{-4.2mm}
\end{figure}

Based on our observations and analysis, we have three critical findings, which are seriously overlooked.
\textbf{First}, \textit{the contribution of image and text modality differs in diverse real-world scenarios.} We observe that previous methods usually prefer to
returning results similar to the query image modality, which fails when the intended image similar to the target does not exist.
This is mainly due to that previous approaches depend heavily on the image modality but underestimate the text modality. Fig.~\ref{fig1} (a) fits this scenario, where the query image is very similar to the target image.
However, in real-world scenarios and the existing datasets, the mixed-modal query is often dominated by the text modality, that is, the similarity between the query image and the target image is very low.
 For example, as shown in Fig.~\ref{fig1} (b), only query text and the concept of \textit{shirt} in query image are conducive to retrieval, while most visual information is redundant or even detrimental. We conduct an exploratory experiment with different modality queries individually in Fig.~\ref{fig2} (a), which verifies modality importance disparity for different categories.
\textbf{Second}, \textit{the widely used datasets \cite{Fashion200k2017, shoes2018, wu2021fashioniq} collected from web are inherently noisy in labeling intention description with text and full of data bias}. This is due to people from different states describe objects and concepts in distinct manners. We conduct an intriguing experiment in Table \ref{tab_text}, in which different text descriptions (templates) show incredibly distinct performances. We observe that the previous SOTA TIRG model~\cite{vo2019composing} shows significantly limited performance. This is due to noisy supervision from different templates and existing training objectives often overfit the noisy data. The relevance between the visual appearance and the text description can thus vary across diverse scenarios.
\textbf{Last but not least}, \textit{the inherent modality gaps affecting the synergistic understanding of multiple modalities are overlooked in training objectives}. Actually, the image-text modality gap makes feature combination challenging, while the mixed-modality gap makes similarity learning difficult.

To tackle the inherent modality importance disparity problem, 
an intuitive idea is directly assigning different weights for different modalities. We conduct an experiment as shown in Fig. \ref{fig2} (b), by following \cite{PL2022}, which uses direct weighting of image-only and text-only models. We find that direct weighting hardly improves the concatenation of two modality features. 
 This is due to that this naive weighting does not delve deeper into the modality features, easily amplifying the error information implied in the dominant modality. Therefore, we suggest editing and purifying the features of different modalities to remove potential error information and dynamically weighting modalities. To tackle the inherent data bias and labeling noise, it is intuitive to enrich the training set with more diverse image and text descriptions, which, however, is labor-intensive. We suggest relaxing the noisy hard labels and exploring dynamic soft similarity labels to fully mine valuable information. Table \ref{tab_text} shows the potential of our proposal. 
Additionally, in training objectives, previous works focus on similarity learning but overlook modality gaps. We propose to utilize large models (e.g., CLIP~\cite{2021Learning}) and enforce mutual enhancement training to naturally bridge modality gaps.

\begin{figure*}[t]
\centering
\includegraphics[width=0.88\textwidth]{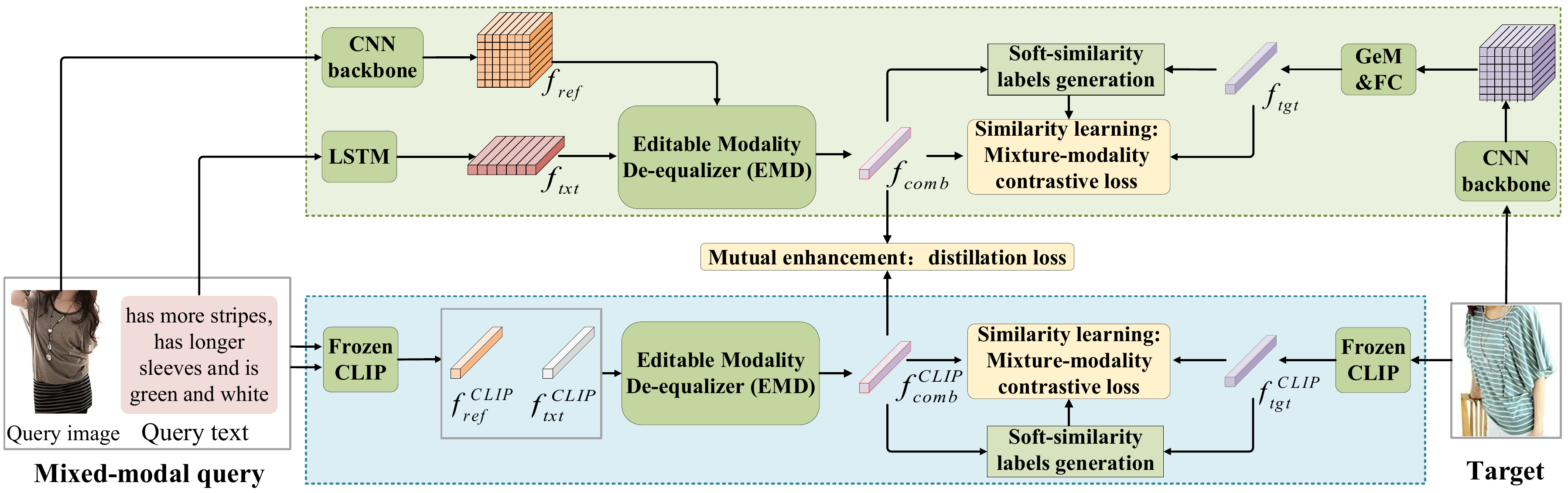}
\vspace{-1.2mm}
\caption{Overview of Dynamic Weighted Combiner (DWC), consisting of two mutually-enhanced streams. Each contains 3 merits: 1) Editable Modality De-equalizer (EMD), 2) dynamic soft-similarity label generator (SSG), and 3) mixed-modality contrastive loss.}
\vspace{-3.2mm}
\label{fig3}
\end{figure*}

Based on above critical findings and motivations, we propose a Dynamic Weighted Combiner (DWC), composed of an editable modality de-equalizer (EMD), a dynamic soft-similarity label generator (SSG), and a mixture modality-image modality contrastive loss. EMD contains two modality editors and an adaptive weighted combiner to purify modality features and unequally treat different modalities, such that modality importance disparity is amended. SSG aims to relax the biased hard labeling in text description by generating soft-similarity labels, which facilitates full utilization of valuable information in datasets, such that noisy supervision is improved. In order to bridge modality gaps and facilitate similarity learning, we introduce a CLIP-based mutual enhancement module alternately trained by a mixed-modality contrastive loss.
Experiments on Fashion200K, Shoes, and FashionIQ datasets show the outstanding performances.
The main contributions are as follows.
\begin{itemize}
  \item We propose a Dynamic Weighted Combiner (DWC) to solve the inherent modality importance disparity, biased labeling noises, and modality gaps. Fig.~\ref{fig3} depicts the overall architecture.
  \item We introduce an Editable Modality De-equalizer (EMD), a dynamic soft-similarity label generator (SSG), and a CLIP-based mixed-modality contrastive training loss to meet the above challenges.
\end{itemize}
\begin{table}[t]
\renewcommand\arraystretch{0.9}
\vspace{-1.2mm}
\caption{Impact of different text descriptions on MMIR performance, which unveils the biased labeling noise in the Fashion200k datasets.}
\centering
\resizebox{\linewidth}{!}{
\begin{tabular}{l|cc|cc}
\hline
\multirow{2}{*}{Text description}& \multicolumn{2}{c|}{TIRG \cite{vo2019composing}} & \multicolumn{2}{c}{Our model}\\
&R@1&R@10  &R@1&R@10\\
\hline
Template 1 \shortcite{vo2019composing} &14.10&42.50 &18.24&48.69\\
Template 2 \shortcite{anwaar2021compositional} &18.10 &52.40 &33.95&60.83\\
Template 3 (ours) &24.50 &49.60  &36.49&63.58\\
\hline
\end{tabular}}
\vspace{-3.2mm}
\label{tab_text}
\end{table}

\section{2. Related Work}
\textbf{Mixed-Modal Image Retrieval (MMIR)} aims to incorporate a query image and text describing the users' intentions to navigate the visual search. Previous approaches for MMIR can be categorized into two types. First, \cite{vo2019composing, JPM2021, Hou2021ICCV, CLVC-Net2021, AACL2022} focus on designing complex components for multi-modal fusion of text and image queries. \cite{vo2019composing} first propose a TIRG network by feature composition of image and text features. \cite{chen2020VAL} propose a VAL framework to fuse image and text features via attention learning at varying representation depths. \cite{CLVC-Net2021} propose a CLVC-Net, which combines local-wise and global-wise composition modules. \cite{GA2022} propose a plugged-and-played gradient augmentation (GA) module based regularization approach to improve the generalization of MMIR models.
Second, \cite{2018Text, tautkute2021i} focus on generating images similar to the target. \cite{tautkute2021i} propose a SynthTripletGAN for image retrieval with synthetic query expansion.

\textbf{Vision-and-Language Pre-training (VLP)} \cite{2021Learning, Vinvl2021, ALBEF2021, METER2022} aims to learn multi-modal representations from large-scale image-text pairs, which has proven to be highly effective on various downstream tasks, such as Visual Question Answering (VQA), Natural Language for Visual Reasoning (NLVR) and Visual Grounding (VG). Recently, VLP models attract attention to solve the MMIR problem. \cite{PL2022} propose a three-stage progressive learning method based on CLIP \cite{2021Learning} to acquire complex knowledge progressively, and fully exploit the open-domain and open-format resources.
\cite{CWCLIP2021, CWCLIP2022, CWCLIP2022w} propose a fine-tuning scheme for conditioned image retrieval using CLIP-based features. \cite{FashionVLP2022} propose a vision-language transformer-based model, FashionVLP, that brings the prior knowledge contained in large image-text corpora to the domain of fashion image retrieval, and combines visual information from multiple levels of context to effectively capture fashion-related information.
However, most previous approaches usually achieve limited performance due to the inherent modality importance disparity and biased labeling noises in datasets.
Based on our findings, we propose a Dynamic Weighted Combiner (DWC) to solve these overlooked problems in MMIR.
\section{3. Dynamic Weighted Combiner}

\subsection{3.1. Problem Definition and Model Architecture}
In MMIR, given a mixed-modal query involving a query image $I_r$  and a text $T$, the objective is to learn image-text combination features
to retrieve the target image $I_t$. In other word, given an input pair $(I_r, T)$, we aim to learn mixed-modal query features $f_{comb} \!= \!\mathcal{C}(I_r, T; \Theta)$ that are well-aligned with the target image feature $f_{tgt}\!=\!\mathcal{F}(I_t; \Theta)$ by maximizing their similarity as,
\begin{equation}
\max_\mathbf{\Theta} \quad \kappa (\mathcal{C}(I_r, T; \Theta),\mathcal{F}(I_t; \Theta)),
\label{eq1}
\end{equation}
where $\mathcal{C}(\cdot)$ and $\mathcal{F}(\cdot)$ denote the mixed-modal feature composer and the image feature extractor, respectively. $\kappa(\cdot, \cdot)$ denotes the similarity kernel, implemented as dot product. $\Theta$ denotes the learnable parameters.
\begin{figure*}[t]
\centering
\includegraphics[width=0.88\textwidth]{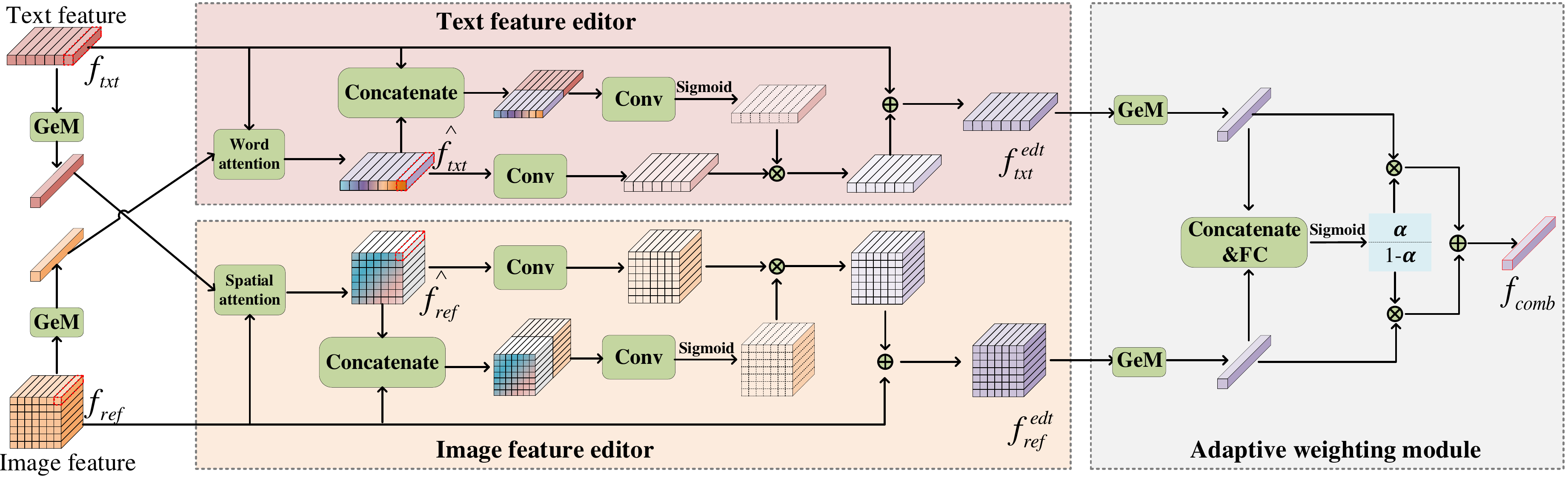}
\caption{Overview of our Editable Modality De-equalizer (EMD) module, which contains an image feature editor, a text feature editor, and an adaptive weighted combiner.  $\bigotimes$ and $\bigoplus$ stand for the Hadamard product and element-wise addition, respectively.}
\label{fig4}
\vspace{-3.2mm}
\end{figure*}

The proposed Dynamic Weighted Combiner (DWC) is illustrated in Fig.~\ref{fig3}, which consists of two mutually-enhanced streams. Each stream is basically composed of four parts: (1) feature encoder to extract the visual and text feature, (2) Editable Modality De-equalizer (EMD) to edit and purify the modality features and unequally combine them with different contributions, (3) dynamic soft-similarity labels generator (SSG) to improve noisy supervision, and (4) mixed-modality contrastive loss to reduce the modality gaps and facilitate similarity learning. The key ingredients are elaborated as follows. 

\subsection{3.2. Feature Encoder}
As is shown in Fig.~\ref{fig3}, the first stream adopts CNN and LSTM as the backbone to train the image encoder and text encoder, respectively. $f_{ref} \in\mathbb{R}^{C\times H\times W}$ and $f_{tgt} \in\mathbb{R}^{C\times H\times W}$ denote the feature maps of the query image and target image extracted by CNN, respectively, where $C\times H\times W$ is the shape of the feature maps.
$f_{txt}\in\mathbb{R}^{D \times L}$ represents the text feature extracted by LSTM, where $D\times L$ is the shape of the text feature and $L$ is the length of the text (i.e., the number of words in the text).
To exploit the prior knowledge of large-scale web data and reduce the modality gap, 
the other stream adopts the large CLIP model as the feature encoder, pre-trained on a large-scale dataset with 400 million image-text pairs scraped from the web. 
$f_{ref}^{CLIP}$, $f_{tgt}^{CLIP}$ and $f_{txt}^{CLIP}$ denote the feature vectors of the query image, target image and query text extracted by pre-trained CLIP, respectively. For convenience, we name the two streams as CNN stream and CLIP stream.

\subsection{3.3. Editable Modality De-equalizer}\label{emd}
Since the contributions of image and text modalities are different in diverse real-world scenarios, an intuitive idea is assigning different weights. However, as discussed in the introduction, direct weighting will unavoidably amplify the error information in the dominant modality and lead to adverse performance. We therefore propose to purify the image and text features before assigning weights to reduce the interference of redundant error information, which gives rise to the Editable Modality De-equalizer (EMD), formalized by two functionally symmetrical modality-feature editors and an adaptive weighted combiner, as shown in Fig.~\ref{fig4}.

\textbf{Image feature editor.}
To purify the image features, using image features as a template, we regard the image feature as $H\!\times \!W$ features of different spatial entities and propose a cross-modal spatial attention mechanism to assign different weights for these features. Specifically, we first get a text representation vector by the GeM pooling layer \cite{GeM2019}. Then the spatial attention $A^{sp}$ is formulated as
\begin{equation}
\begin{aligned}
&A^{sp} = Reshape(Softmax(S^{sp})),\\
&S_{i,j}^{sp} = \kappa({f}_{ref}(:,i,j), GeM(f_{txt})),
\end{aligned}
\label{eq2}
\end{equation}
where $A^{sp} \in\mathbb{R}^{1 \times H \times W}$ and $S_{i,j}^{sp}$ indicate the similarity between the image feature of the $(i,j)^{th}$ spatial entity and the text feature. Accordingly, we can derive a coarse-modified image feature:
\begin{equation}
\widehat{f}_{ref} = A^{sp} \otimes {f}_{ref},
\label{eq3}
\end{equation}
In order to further refine the image feature, we transform the feature map through element-level global attention, which can be formulated as
\begin{equation}
A_{ref}^{gl} = Sigmoid (Conv([\widehat{f}_{ref},f_{ref}])),
\label{eq4}
\end{equation}
where $A_{ref}^{gl} \in\mathbb{R}^{C \times H \times W}$. $Conv(\cdot)$ and $[\cdot,\cdot]$ denote $1\times 1$ convolution layer and concatenation, respectively. Ultimately, the purified image features can be obtained as
\begin{equation}
f_{ref}^{edt} = (A_{ref}^{gl}\otimes Conv(\widehat{f}_{ref})) \oplus {f}_{ref},
\label{eq5}
\end{equation}

\textbf{Text feature editor.} Similarly, to eliminate the error information 
from the text, we introduce a text feature editor. Using text features as a template, we propose a cross-modal word attention mechanism to assign weights for different words. Specifically, the word attention $A^{w}$ is formulated as
\begin{equation}
\begin{aligned}
&A^{w} = Reshape(Softmax(S^w)),\\
&S_{l}^w = \kappa({f}_{txt}(:l), GeM(f_{ref})),
\end{aligned}
\label{eq6}
\end{equation}
where $A^{w} \in\mathbb{R}^{1 \times L}$ and $S_{l}^w$ indicate the similarity between the text feature of the $l^{th}$ word and the image feature. Accordingly, we can derive a coarsely-modified text feature:
\begin{equation}
\widehat{f}_{txt} = A^{w} \otimes {f}_{txt},
\label{eq7}
\end{equation}
In order to further refine the text feature, we transform the feature through element-level global attention, which can be formulated as
\begin{equation}
A_{txt}^{gl} = Sigmoid (Conv([\widehat{f}_{txt},f_{txt}])),
\label{eq8}
\end{equation}
where $A_{txt}^{gl} \in\mathbb{R}^{D \times L}$. Ultimately, the purified text features can be obtained as
\begin{equation}
f_{txt}^{edt} = (A_{txt}^{gl}\otimes Conv(\widehat{f}_{txt})) \oplus {f}_{txt},
\label{eq9}
\end{equation}

\textbf{Adaptive weighting module.} After obtaining the refined features, we introduce an adaptive weighting module to assign different modality weights for the query image and text according to their contributions. Specifically, we use $\alpha$ to represent the importance of the image and $1-\alpha$, the importance of the text. $\alpha$ is computed as
\begin{equation}
\alpha =Sigmoid( FC([GeM(f_{ref}^{edt}),GeM(f_{txt}^{edt})])),
\label{eq10}
\end{equation}
where $FC$ denotes fully-connected layers. The final combination feature is formulated as
\begin{equation}
f_{comb} = \alpha \cdot f_{ref}^{edt} +(1-\alpha) \cdot f_{txt}^{edt}.
\label{eq11}
\end{equation}

Intuitively, the feature editors purify the feature maps in two steps, i.e., coarsely modifying via spatial/word attention and refining via global attention. For the CLIP stream, to reduce the computation cost, we adopt lightweight global attention to edit the CLIP features only in one step (see \emph{Supplementary Material}).

\subsection{3.4. Soft-Similarity Labels Generation}\label{sl}
\begin{figure}[t]
\centering
\includegraphics[width=0.42\textwidth]{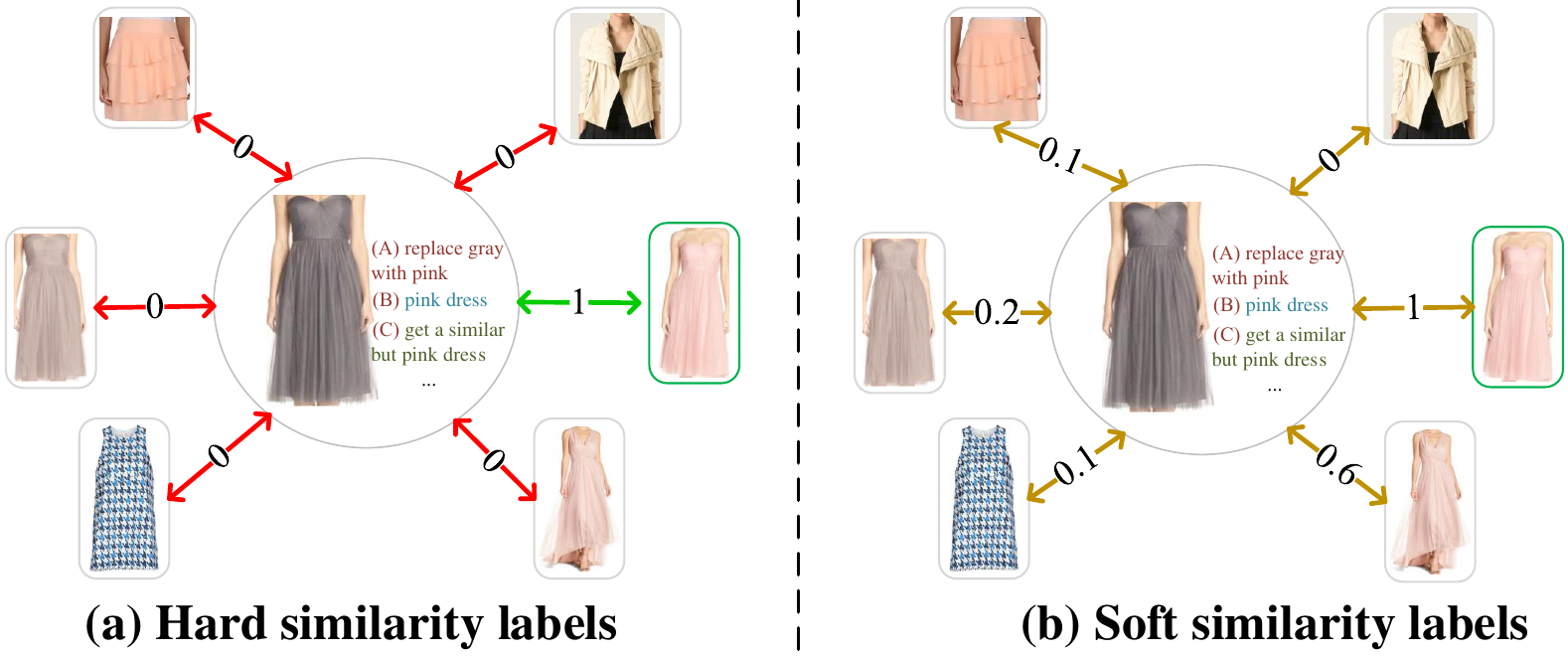}
\caption{Basic idea of the proposed soft-similarity labels.}
\label{fig5}
\vspace{-3.2mm}
\end{figure}

The web datasets for MMIR tend to be seriously prejudiced and noisy in labeling text modality descriptions, because people from different states describe objects and concepts in distinct manners. Most previous approaches overfit the biased noisy data during training, as discussed in the introduction, because the original one-hot (hard) similarity labels are usually imprecise. For example, as shown in Fig. \ref{fig5} (a), the hard similarity label for positive sample, i.e., $(I_r, T)$ and $I_t$ in the same triplet $<I_r, T, I_t>$, otherwise is $0$ (i.e., negative sample). In fact, some negative samples can also serve as the target for some users, but re-labeling is undoubtedly a labor-intensive task.

Since different language descriptions result in different similarity scores between the query and the target, we propose to relax the hard-similarity labels with soft-similarity labels, to avoid overfitting and simultaneously fully mine the valuable information in datasets, which gives rise to the soft-similarity labels generator (SSG).
Specifically, we generate nonzero dynamic soft-similarity labels on-the-fly for negative pairs only. The soft-similarity label between the $i^{th}$ query and $j^{th}$ target in a mini-batch is generated by
\begin{equation}
y_{ij}^s = \left\{
\begin{array}{lcl}
1 ,&if \ \ i=j,\\
\frac{exp(\kappa (f_{tgt}^i,f_{tgt}^j)/ \tau)}{\sum_{j=1}^{B} exp(\kappa (f_{tgt}^i,f_{tgt}^j) /\tau)}, & if \ \ i\neq j,\\
\end{array}\right.
\label{eq14}
\end{equation}
where $\tau$ is a temperature factor. Finally, let $\mathbf{Y}^s = \{y_{ij}^s\}_{i=1,j=1}^{B,B} \in[0,1]$ denote all soft-similarity labels, and $B$ is the batch size. A toy example of the soft-similarity labels is shown in Fig. \ref{fig5} (b).
\subsection{3.5. Mixed-modality Contrastive Loss}\label{micl}
To facilitate similarity learning, we introduce a mixed-modality contrastive loss between the combined features and the target image features based on the soft-similarity labels. 
Specifically, for each mixed-modal query and target, we first compute the softmax-normalized similarity as:
\begin{equation}
p_{ij} =  \frac{exp(\kappa (f_{comb}^i,f_{tgt}^j)/ \tau)}{\sum_{j=1}^{B} exp(\kappa (f_{comb}^i,f_{tgt}^j) /\tau)},
\label{eq15}
\end{equation}
where 
$\mathbf{P}^s = \{p_{ij}\}_{i=1,j=1}^{B,B}$ represents the predicted probability.  Let $\mathbf{Y}^h$ denote the noisy hard ground-truth similarity. The cross-entropy based mixed-modality contrastive loss with soft-similarity guidance is written as
\begin{equation}
\begin{aligned}
&\mathcal{L}_{mmc} = \lambda \mathbb{E}[H(\mathbf{Y}^h, \mathbf{P})]+ (1-\lambda) \mathbb{E}[H(\mathbf{Y}^s, \mathbf{P})]\\
&=- \frac{\lambda}{B}\sum_{i=1}^{B} \log \frac{exp(\kappa (f_{comb}^i,f_{tgt}^i)/ \tau)}{\sum_{j=1}^{B} exp(\kappa (f_{comb}^i,f_{tgt}^j) /\tau)}\\
&-\frac{(1-\lambda)}{B^2}\sum_{i=1}^{B} \sum_{j=1}^{B}\frac{exp(\kappa (f_{tgt}^i,f_{tgt}^j)/ \tau)}{\sum_{j=1}^{B} exp(\kappa (f_{tgt}^i,f_{tgt}^j) /\tau)} \\
&\cdot \log \frac{exp(\kappa (f_{comb}^i,f_{tgt}^j)/ \tau)}{\sum_{j=1}^{B} exp(\kappa (f_{comb}^i,f_{tgt}^j) /\tau)},
\end{aligned}
\label{eq16}
\end{equation}
where $\lambda \in (0,1)$ is a trade-off parameter and $H(\cdot)$ is the cross-entropy loss.

\subsection{3.6. Model Training}\label{train}
Inspired by the idea of mutual learning \cite{MUL2018}, we introduce a mutual enhancement strategy to train two streams alternately, between which the knowledge is shared via a vanilla KL-divergence based distillation loss. Specifically, we first compute the similarity between the $i^{th}$ query and $j^{th}$ target of both CNN and CLIP streams, resp. as follows
\begin{equation}
p_{ij}^z =  \frac{exp(\kappa ((f_{comb}^{i})^z,(f_{tgt}^{j})^z)/ \tau)}{\sum_{j=1}^{B} exp(\kappa ((f_{comb}^i)^z,(f_{tgt}^j)^z) /\tau)},
\label{eq17}
\end{equation}
where $z\in$\{CNN, CLIP\}. In this way, we can get the similarity distribution of the $i^{th}$ query as $p_{i}^z \!= \![p_{i1}^z, p_{i2}^z, ..., p_{iB}^z]$. Then the distillation loss is
\begin{equation}
\mathcal{L}_{d}^z  = \! \frac{1}{B}\sum_{i=1}^{B} D_{KL}(\mathbf{p}_{i}^{\bar{z}} ||p_{i}^z)
= \!\frac{1}{B^2}\sum_{i=1}^{B}\sum_{i=1}^{B}p_{ij}^{\bar{z}} \log \frac{p_{ij}^{\bar{z}}}{p_{ij}^z}.
\label{eq18}
\end{equation}
Taking the optimization of the CNN stream as an example, we use $\mathcal{L}_{d}^{CNN}$ to transfer the knowledge from the CLIP stream to the CNN stream, and there is
\begin{equation}
\mathcal{L}_{d}^{CNN}  = \frac{1}{B^2}\sum_{i=1}^{B}\sum_{i=1}^{B}p_{ij}^{CLIP} \log \frac{p_{ij}^{CLIP}}{p_{ij}^{CNN}}.
\label{eq19}
\end{equation}
The total loss of the CNN stream is formulated as:
\begin{equation}
\mathcal{L}_{all}^{CNN}  = \mathcal{L}_{mmc}^{CNN} + \mathcal{L}_{d}^{CNN}.
\label{eq20}
\end{equation}
Notably, the total loss of the CLIP stream can be derived similarly. In the training phase, we alternately optimize the two streams to achieve mutual enhancement. Finally, the combination features of the two streams are integrated to evaluate the similarity and rank the gallery images.

\textbf{Mutual information maximization perspective.} Minimizing the first term of $\mathcal{L}_{mmc}$ (Eq.~\ref{eq16}) can be seen as maximizing the lower bound on the mutual information (MI) between the mixed-modal query and the target, i.e., maximizing a symmetric version of InfoNCE \cite{InfoNCE2018}. Therefore, the proposed $\mathcal{L}_{mmc}$ can not only learn the mixed-modal similarities for MMIR, but also reduce the modality gaps.

\section{4. Experiments}

To evaluate our model, we chose three real-world datasets: Fashion200K \cite{Fashion200k2017}, Shoes \cite{shoes2018}, and FashionIQ \cite{wu2021fashioniq}. The datasets and implementation details are described in the \textit{supplementary material}.
We compare our DWC with many SOTA MMIR methods, such as \textbf{TIRG} \cite{vo2019composing}, \textbf{JAMMAL} \cite{2020Joint}, \textbf{LBF} \cite{ho2020CQIR}, \textbf{JVSM} \cite{chen2020JVSM}, \textbf{SynthTripletGAN} \cite{tautkute2021i}, \textbf{VAL} \cite{chen2020VAL}, \textbf{DCNet}\cite{DCNet2021}, \textbf{JPM} \cite{JPM2021}, \textbf{DATIR} \cite{DATIR2021}, \textbf{ComposeAE} \cite{anwaar2021compositional}, \textbf{CoSMo} \cite{lee2021CoSMo}, \textbf{CLVC-Net} \cite{CLVC-Net2021}, \textbf{ARTEMIS} \cite{ARTEMIS2022}, \textbf{SAC} \cite{SAC2022}, \textbf{GA} \cite{GA2022},
\emph{\textbf{CIRPLANT}} \cite{CT2021}, \textbf{\emph{Combiner w/ CLIP}} \cite{CWCLIP2022}, and \emph{\textbf{FashionVLP}} \cite{FashionVLP2022}, where the methods in italic are based on VLP models.

\begin{table*}
\tiny
\renewcommand\arraystretch{0.4}
\caption{Interactive image retrieval performance (\%) on the FashionIQ dataset. The best results are in bold.}
\vspace{-2mm}
\centering
\setlength{\tabcolsep}{3mm}{
\resizebox{\linewidth}{!}{
\begin{tabular}{l|cc|cc|cc|ccc}
\hline
\multirow{2}{*}{Method}& \multicolumn{2}{c|}{Dress} & \multicolumn{2}{c|}{Shirt} & \multicolumn{2}{c|}{Toptee} & \multicolumn{3}{c}{Average} \\
&R@10&R@50 &R@10&R@50 &R@10&R@50 &R@10&R@50 &Mean\\
\hline
TIRG &14.87&34.66&18.26&37.89&19.08&39.68&17.40&37.41&27.41\\
VAL &22.53&44.00&22.38&44.15&27.53&51.68&24.15&46.61&35.38\\
ComposeAE &14.03&35.10&13.88&34.59&15.80&39.26&14.57&36.32&25.44\\
JVSM &10.70&25.90&12.00&27.10&13.00&26.90&11.90&26.63&19.27\\
SynthTripletGAN &22.60&45.10&20.50&44.08&28.01&52.10&23.70&47.09&35.40\\
CoSMo &25.64&50.30&24.90&49.18&29.21&57.46&26.58&52.31&39.45\\
JPM &21.38&45.15&22.81&45.18&27.78&51.70&23.99&47.34&35.67\\
DATIR &21.90&43.80&21.90&43.70&27.20&51.60&23.67&46.37&35.02\\
CLVC-Net &29.85&56.47&28.75&54.76&33.50&64.00&30.70&28.41&44.56\\
ARTEMIS &27.34&44.18&21.05&49.87&24.91&48.59&24.43&47.55&35.99\\
SAC &26.13&52.10&26.20&50.93&31.16&59.05&27.83&54.03&40.93\\
\hline
CIRPLANT &17.45&40.41&17.53&38.81&21.64&45.38&18.87&41.53&30.20\\
Combiner w/ CLIP &26.82&51.31&31.80&53.38&33.40&57.01&30.67&53.90&42.29\\
FashionVLP &32.42&\textbf{60.29}&31.89&58.44&38.51&\textbf{68.79}&34.27&\textbf{62.51}&48.39\\
\hline
DWC&\textbf{32.67}&57.96&\textbf{35.53}&\textbf{60.11}&\textbf{40.13}&66.09&\textbf{36.11}&61.39&\textbf{48.75}\\
\hline
\end{tabular}}}
\vspace{-2mm}
\label{tab3}
\end{table*}

\begin{table}[t]
\renewcommand\arraystretch{0.5}
\centering
\caption{Performance Comparison (\%) on Fashion200k.}
\vspace{-2mm}
\setlength{\tabcolsep}{4mm}{
\resizebox{\linewidth}{!}{

\begin{tabular}{l|ccc|c}
\hline
Method  &R@1&R@10&R@50&Average\\
\hline
TIRG &14.10&42.50&63.80&40.13\\
JVSM &19.00	&52.10	&70.00	&47.03\\
JAMMAL &17.30	&45.30	&65.70	&42.77\\
LBF   &17.80	&48.40	&68.50	&44.90\\
VAL	&22.90	&50.80	&72.70	&48.80\\
DCNet &-- &46.90	&67.60   &--\\
JPM &19.80	&46.50	&66.60	&44.30\\
DATIR &21.50	&48.80	&71.60	&47.30\\
ComposeAE  &22.80	&55.30	&73.40	&50.50\\
CoSMo  &23.30	&50.40	&69.30	&47.67\\
CLVC-Net &22.60	&53.00	&72.20	&49.27\\
ARTEMIS  &21.50	&51.10	&70.50	&47.70\\
GA &24.00	&57.20	&75.70	&52.30\\
\hline
Combiner w/ CLIP &20.56&52.07&71.35&47.99\\
FashionVLP  &-- &49.90	&70.50   &--\\
\hline
DWC &\textbf{36.49}	&\textbf{63.58}	&\textbf{79.02} &\textbf{59.70}\\
\hline
\end{tabular}}}
\label{tab1}
\end{table}
\subsection{4.1. Experimental Results}

\begin{table}[!ht]
\renewcommand\arraystretch{0.5}
\centering
\caption{Performance Comparison (\%) on Shoes.}
\vspace{-2mm}
\setlength{\tabcolsep}{4mm}{
\resizebox{\linewidth}{!}{
\begin{tabular}{l|ccc|c}
\hline
Method  &R@1&R@10&R@50&Average\\
\hline
TIRG &12.60&45.45&69.39&42.48\\
VAL	&17.18&51.52&75.83&48.18\\
SynthTripletGAN &--&47.6&73.6&--\\
ComposeAE &4.37&19.36&47.58&23.77\\
CoSMo &16.72&48.36&75.64&46.91\\
DATIR &17.20&51.10&75.60&47.97\\
DCNet &-- &53.8&79.3 &--\\
CLVC-Net &17.60&54.40&79.50&50.50\\
ARTEMIS &17.6&51.05&76.85&48.50\\
SAC &18.11&52.41&75.42&48.65\\
\hline
Combiner w/ CLIP &8.12&33.28&62.42&34.61\\
FashionVLP &-- &49.08&77.32  &--\\
\hline
DWC &\textbf{18.94}&\textbf{55.55}&\textbf{80.19}&\textbf{51.56}\\
\hline
\end{tabular}}}
\label{tab2}
\end{table}

\textbf{Quantitative Results.} Table \ref{tab3} shows comparisons with existing methods on the FashionIQ dataset. We observe that the performance improvement of DWC over the second-best method is 0.25\%, 3.64\% and 1.98\% on R@10 for three subsets, i.e., dress, shirt, and top-tee, respectively. Our DWC is slightly lower than FashionVLP in R@50, a more complex model equipped with auxiliary modules such as object detection module and landmark module. Table \ref{tab1} shows our comparison with existing methods on the Fashion200k dataset. As can be seen, our model demonstrates compelling results compared to all other alternatives. The performance improvement of DWC over the second-best method is 12.49\% and 7.4\% on R@1 and the average, respectively.
Table \ref{tab2} shows comparisons with existing methods on the Shoes dataset. Our DWC model still achieves the best performance with  improvements of 0.83\% and 1.06\% on R@1 and the average, respectively.

%
%


\begin{figure}[h]
\centering
 \centerline{\includegraphics[width=0.39\textwidth]{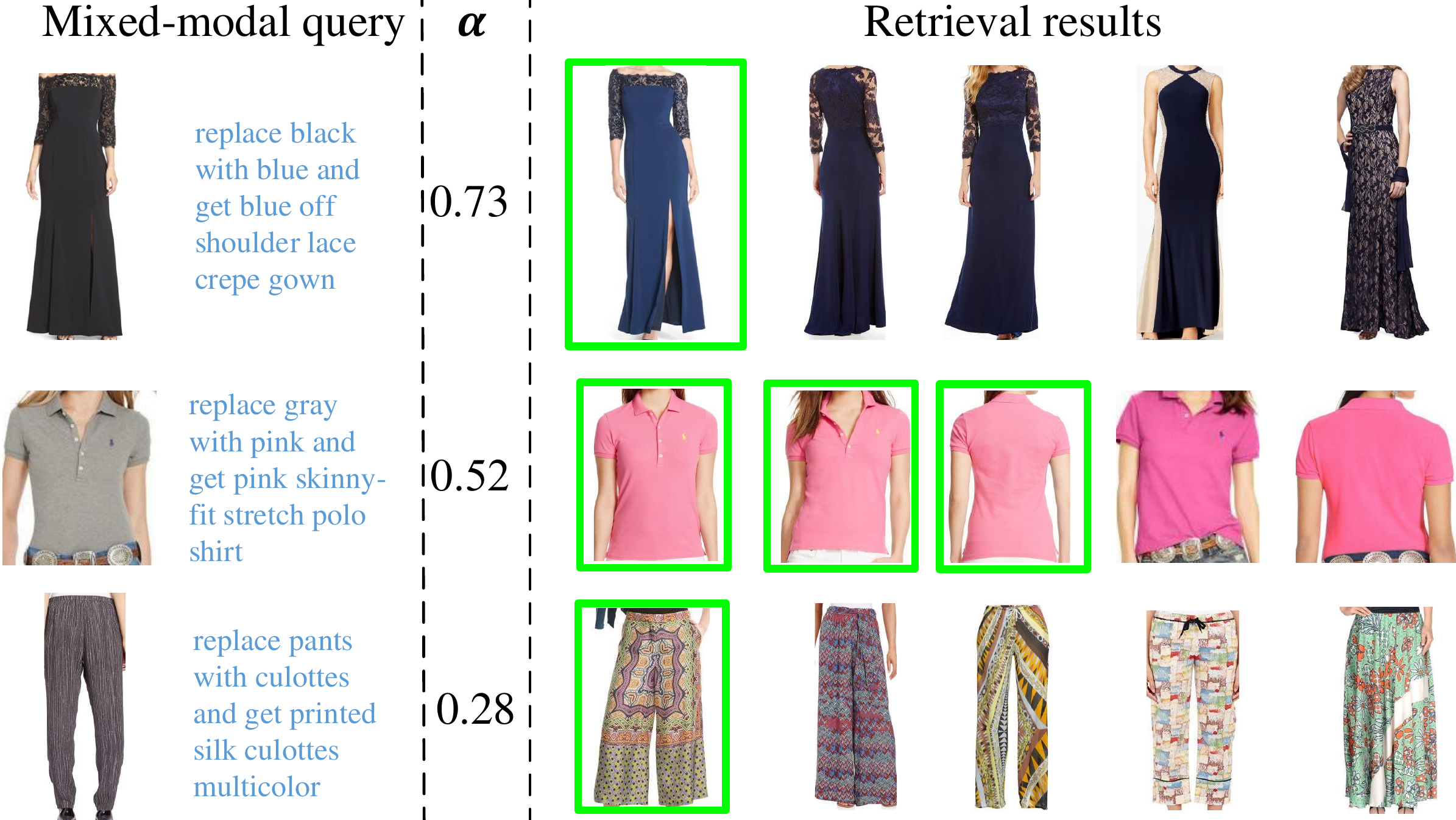}}
\caption{Qualitative results on Fashion200k. $\alpha$ is the importance of image modality. Ground-truths are shown in green. }
\label{fig200k}
\vspace{-1.2mm}
\end{figure}
\begin{figure}[h]
\vspace{-1.2mm}
\centering
 \centerline{\includegraphics[width=0.41\textwidth]{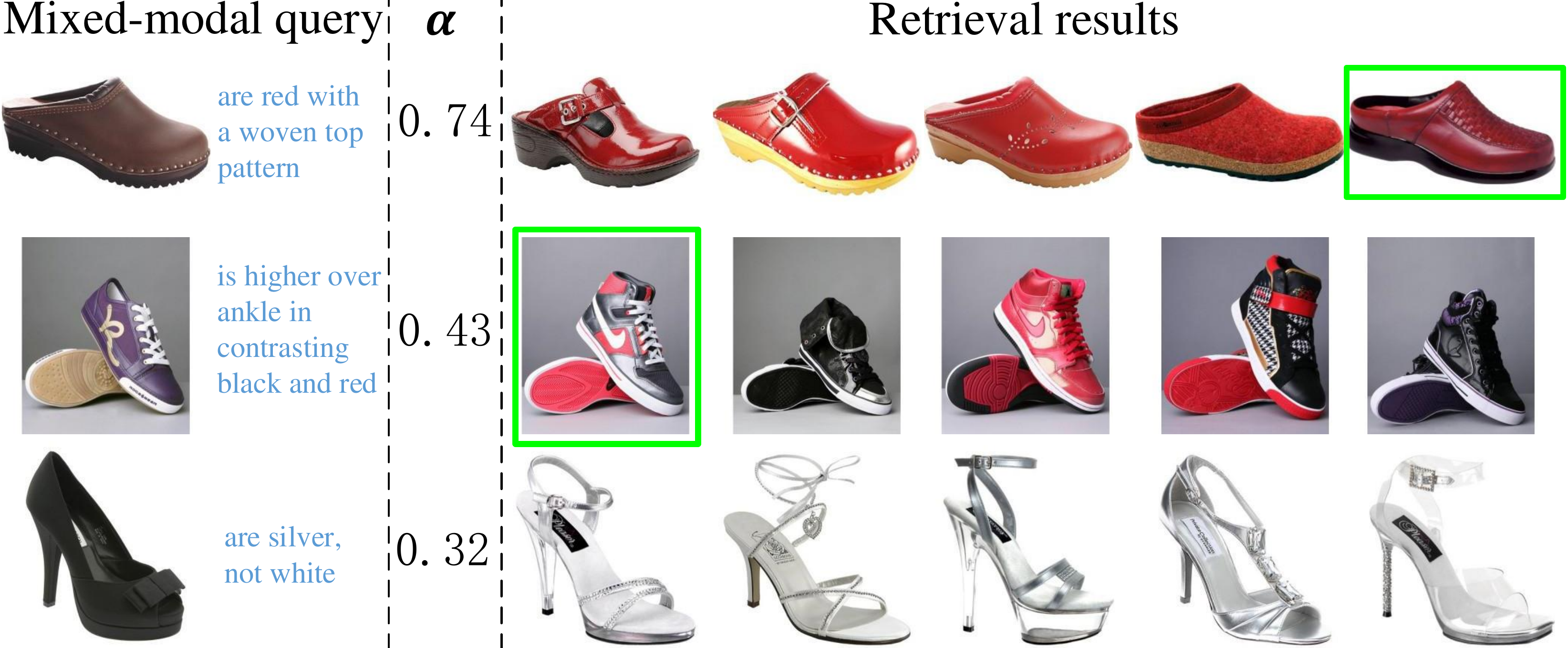}}
\caption{Qualitative results on the Shoes dataset.
}
\vspace{-1.2mm}
\label{figshoes}

\end{figure}
 \begin{figure}[!h]
 \vspace{-1.2mm}
\centering
 \centerline{\includegraphics[width=0.41\textwidth]{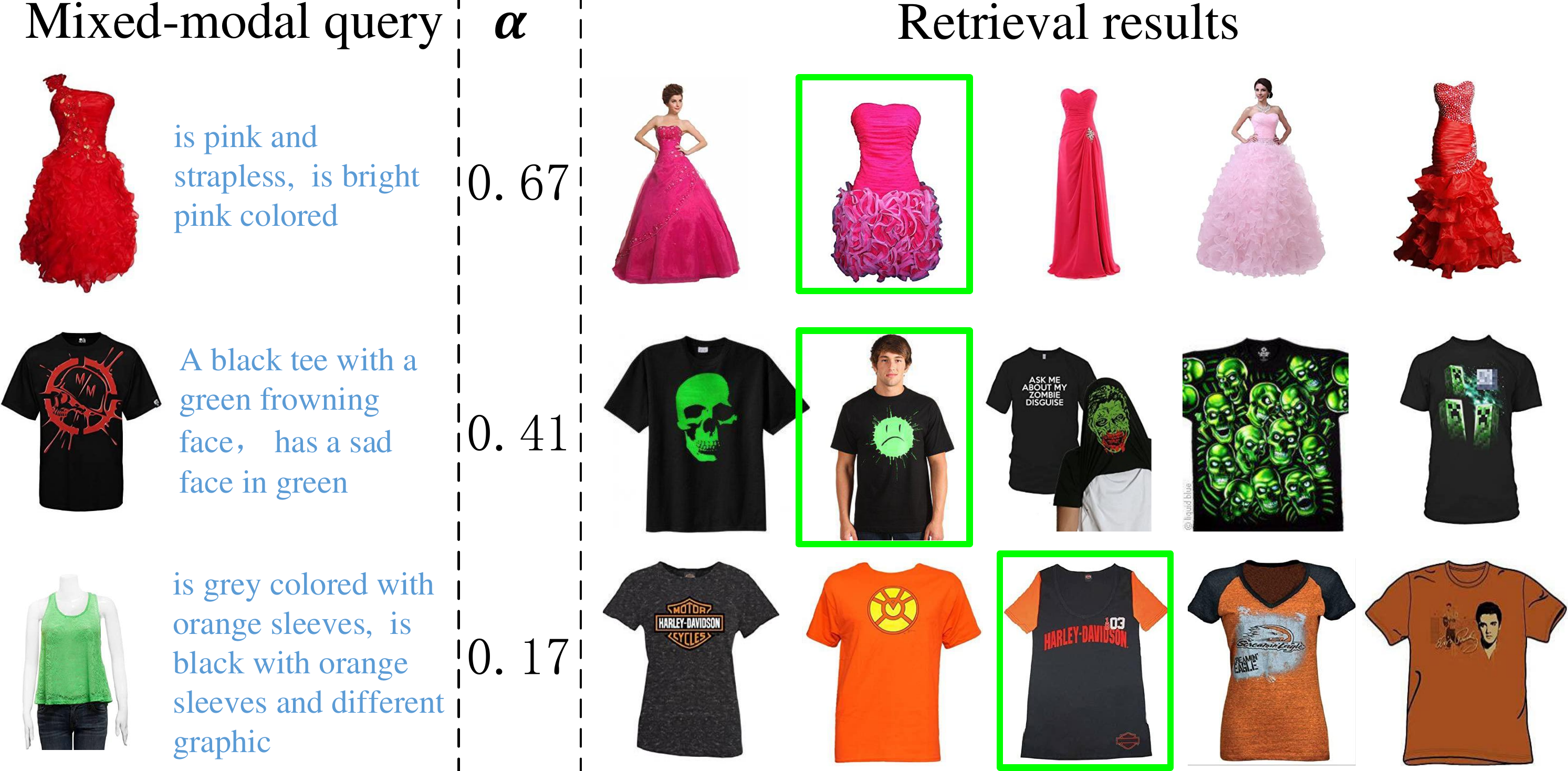}}
\caption{Qualitative results on the FashionIQ dataset.
}
\vspace{-1.2mm}
\label{figiq}
\vspace{-3.2mm}
\end{figure}

\begin{figure*}
\centering
 \centerline{\includegraphics[width=0.9\textwidth]{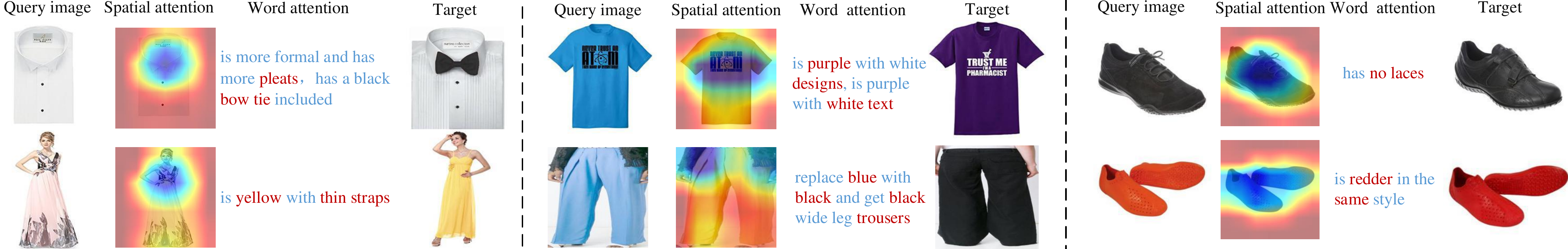}}
\caption{Attention visualization of our model on different datasets. Words with high attention value are in red.}
\vspace{-3.2mm}
\label{attention}
\end{figure*}
\textbf{Qualitative Results and Analysis}.
Fig.~\ref{fig200k} shows our qualitative results on the Fashion200k dataset. Our model can perceive both simple visual attributes like color and complex visual properties like pants and culotte for retrieving candidate images. Fig.~\ref{figshoes} shows our qualitative results on the Shoes dataset. The last row denotes failure cases, where the model incorrectly predicts that the text modality is more important than the image, and thus wrongly returns the candidates only associated with the text. This is mainly due to that the text description is incomplete and unrelated to the query picture. It is difficult to predict the importance of modalities when the text is independent of the query image. Fig.~\ref{figiq} shows our qualitative results on the FashionIQ datasets. Results show that our model can capture such diverse concepts and retrieve good candidate images. Fig.~\ref{attention} provides the attention visualizations, which show the proposed spatial attention and word attention are effective.

\subsection{4.2. Ablation Study and Model Analysis}
\textbf{Ablation study of losses.} 
The results are provided in Table \ref{Rtab1}. 
DWC w/o $\mathcal{L}_{mmc-hard}$ and DWC w/o $\mathcal{L}_{mmc-soft}$ represent remove $\mathbb{E}[H(\mathbf{Y}^h,\mathbf{P})]$ and $ \mathbb{E}[H(\mathbf{Y}^s, \mathbf{P})]$ from Eq. (14), respectively. DWC w/o $\mathcal{L}_{d}$ means remove the distillation loss, i.e., Eq. (16).
Results reveal the effectiveness of the losses and mutual enhancement strategy. 
\begin{table}
\renewcommand\arraystretch{1.2}
\caption{Ablation study of losses.}
\vspace{-2mm}
\centering
\setlength{\tabcolsep}{0.8mm}{
\resizebox{\linewidth}{!}{
\begin{tabular}{l|cc|cc|cc}
\hline
\multirow{2}{*}{Method}& \multicolumn{2}{c|}{Dress} & \multicolumn{2}{c|}{Shirt} & \multicolumn{2}{c}{Toptee} \\
&R@10&R@50 &R@10&R@50 &R@10&R@50 \\
\hline
DWC w/o $\mathcal{L}_{mmc-hard}$  &31.82	&57.11	&34.89	&61.19	&39.37	&65.37\\
DWC w/o $\mathcal{L}_{mmc-soft}$  &31.73&57.80&34.99&59.91&39.42&65.63\\
DWC w/o $\mathcal{L}_{d}$ &29.74	&55.82	&32.04	&59.13	&36.21	&65.12\\
DWC&\textbf{32.67}&\textbf{57.96}&\textbf{35.53}&\textbf{60.11}&\textbf{40.13}&\textbf{66.09}\\
\hline
\end{tabular}}}
\label{Rtab1}
\end{table}

\begin{table}[t]
\renewcommand\arraystretch{0.5}
\caption{Ablation study (R@10). }
\vspace{-2mm}
\centering
\setlength{\tabcolsep}{4mm}{
\resizebox{\linewidth}{!}{

\begin{tabular}{l|cccc}
\hline
Method & {Dress} & {Shirt} &{Toptee} & {Shoes} \\
\hline
DWC w/o EMD&23.55&26.40&29.37&50.72\\
DWC w/o SSG&31.73&34.99&39.42&55.44\\
DWC w/o CLIP  &24.79&24.48&29.37&52.32\\
DWC w/o ME &28.01&27.82	&31.00	&53.41\\
\hline
EMD w/o TFE &30.49 &32.34 &36.92 &52.54\\
EMD w/o IFE &28.01 &30.72 &34.12 &51.86\\
EMD w/o FE &27.37 &29.39 &33.61 &49.65\\
\hline
DWC&\textbf{32.67}&\textbf{35.53}&\textbf{40.13}&\textbf{55.55}\\
\hline
\end{tabular}}}
\vspace{-3mm}
\label{tab4}
\end{table}

\begin{table}[t]
\renewcommand\arraystretch{0.5}
\caption{Impact of different combination methods (R@10). }
\vspace{-2mm}
\centering
\setlength{\tabcolsep}{4mm}{
\resizebox{\linewidth}{!}{
\begin{tabular}{l|cccc}
\hline
Method& {Dress} &{Shirt} & {Toptee} & {Shoes} \\
\hline
Image Only&7.83&13.74&10.20&31.92\\
Text Only&18.79&22.52&25.50&15.39\\
Summation&23.55&26.40&29.37&50.72\\
Concatenation&28.31&30.52&35.03&50.50\\
Weighting &28.36&30.30&34.98&50.16\\
Our EMD&\textbf{32.67}&\textbf{35.53}&\textbf{40.13}&\textbf{55.55}\\
\hline
\end{tabular}}}
\label{tab6}
\end{table}

\begin{table}[t]
\renewcommand\arraystretch{0.5}
\caption{Impact of pre-trained CLIP models (R@10). }
\vspace{-2mm}
\centering
\setlength{\tabcolsep}{4mm}{
\resizebox{\linewidth}{!}{
\begin{tabular}{l|cccc}
\hline
CLIP model& {Dress} & {Shirt} &{Toptee} &{Shoes} \\
\hline
RN50&32.67&35.53&40.13&55.55\\
RN101&32.42&35.97&39.93&54.25\\
ViT-B32&31.63&34.99&38.55&\textbf{56.26}\\
ViT-B16&\textbf{33.61}&\textbf{37.09}&\textbf{40.80}&54.92\\
\hline
\end{tabular}}}
\vspace{-3.2mm}
\label{tab5}
\end{table}
\textbf{Analysis of different components}. To demonstrate the contributions from different components in our proposed model, we conduct ablation studies in Table \ref{tab4}, in which \textit{DWC w/o EMD}, \textit{DWC w/o CLIP}, \textit{DWC w/o SSG} and \textit{DWC w/o ME} indicate the variants of our DWC by removing EMD, CLIP stream SSG, and mutual enhancement, respectively. Contribution of different components in EMD is also evaluated, \textit{EMD w/o TFE}, \textit{EMD w/o IFE}, and \textit{EMD w/o FE} are the variants of DWC by removing text feature editor, image feature editor and two feature editors from EMD, respectively. Experiments show that each component plays a significant role in improving the MMIR performance. This further verifies our intuition to meet the challenges of inherent modality importance disparity, biased labeling noises in datasets and modality gaps.


\textbf{Analysis of modality importance disparity.}
The modality importance disparity, including its impact and the effectiveness of our EMD, is presented in Table \ref{tab6}. The first two rows indicate that different modalities play different roles. The third and fourth rows indicate that mixed-modality feature fusion (sum. vs. concat.) can effectively improve the performance. The fifth row shows that naive weighting of two modalities cannot improve modality importance disparity because error of the important modality is also amplified. In contrast, the proposed EMD can significantly improve the performances to a large margin by taking into account the purification of mixed-modality features with an adaptive weighting combiner.

\textbf{Impact of pre-trained CLIP models under different backbones}. We consider four versions of the pre-trained CLIP models in the CLIP stream to conduct experiments. As presented in Table \ref{tab5}, we observe slight differences among them, which, instead, indicate that the modality gap can be largely bridged by a vanilla CLIP model.

Notably, \textbf{the impact of different formulas to generate soft labels} and many other experimental analyses are provided in the \textit{supplementary material}.


\section{5. Conclusion}
We have two critical findings that are seriously overlooked in MMIR community. 1) There exists significant modality importance disparity, leading to degradation of model training. 2) There exists inherent labeling noises and data bias due to diverse text descriptions crawled from web scenarios. Based on the findings, we propose a Dynamic Weighted Combiner (DWC), which includes three merits.
First, we propose an Editable Modality De-equalizer (EMD) by taking into account the contribution disparity between mixed modalities.
Second, we propose a dynamic soft-similarity label generator (SSG) to relax the biased hard labeling and implicitly improve noisy supervision.
Finally, to bridge modality gaps and facilitate similarity learning, we propose a CLIP-based mutual enhancement module alternately trained by a mixed-modality contrastive loss.
Experiments and analysis verify the superiority of our approach.
\section{Acknowledgement}
This work was partially supported by National Key R\&D Program of China (2021YFB3100800), National Natural Science Fund of China (62271090), and Chongqing Natural Science Fund (cstc2021jcyj-jqX0023).


\bibliography{aaai24}

\section{Supplementary Material}
\subsection{1. Details of CLIP Stream in EMD}
\begin{figure}[h]
\centering
\includegraphics[width=0.4\textwidth]{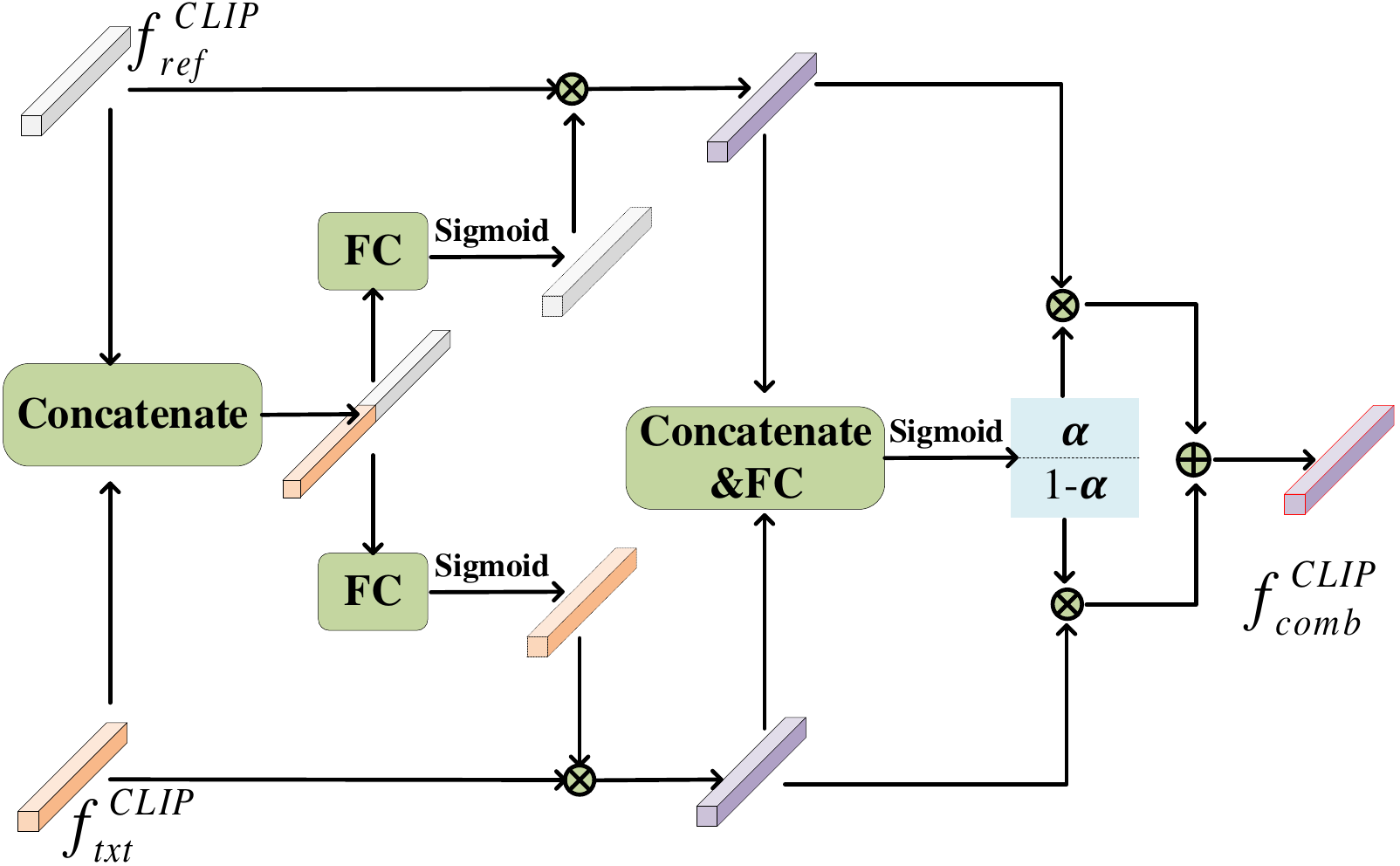}
\caption{Overview of EMD for the CLIP stream.}
\label{Sfig1}
\end{figure}

Due to the space limitation of the main body, we put some details of the CLIP stream in the proposed Editable Modality De-equalize (EMD) module (\textbf{Section 3.3} of the main body paper) in the supplementary material. The details of EMD for the CLIP stream are as follows. For the CLIP stream, to reduce the computation cost, we adopt lightweight global attention to edit the CLIP features only in one step, as shown in Fig.~\ref{Sfig1}. Specifically, the lightweight global attention for CLIP is written as
\begin{equation}
\begin{aligned}
&A_{ref}^{CLIP} =Sigmoid(FC_{ref}([f_{ref}^{CLIP},f_{txt}^{CLIP}]),\\
&A_{txt}^{CLIP} =Sigmoid(FC_{txt}([f_{ref}^{CLIP},f_{txt}^{CLIP}]),
\end{aligned}
\label{Seq1}
\end{equation}
where $A_{ref}^{CLIP}, A_{txt}^{CLIP}\in\mathbb{R}^D$, $FC(\cdot)$ means fully-connected layer and $[\cdot,\cdot]$ means feature concatenation operator. The final combination feature vectors can be represented as
\begin{equation}
f_{comb}^{CLIP} = A_{ref}^{CLIP}\otimes f_{ref}^{CLIP}+ A_{txt}^{CLIP}\otimes f_{txt}^{CLIP}.
\label{Seq2}
\end{equation}

\begin{table*}
\caption{Ablation study of different components in the proposed DWC approach. }
\vspace{-2mm}
\centering
\setlength{\tabcolsep}{5mm}{
\resizebox{\linewidth}{!}{
\begin{tabular}{l|cc|cc|cc|ccc}
\hline
\multirow{2}{*}{Method}& \multicolumn{2}{c|}{Dress} & \multicolumn{2}{c|}{Shirt} & \multicolumn{2}{c|}{Toptee} & \multicolumn{3}{c}{Shoes} \\
&R@10&R@50 &R@10&R@50 &R@10&R@50 &R@1&R@10 &R@50\\
\hline
 DWC w/o EMD&23.55&49.00&26.40&50.29&29.37&55.33&15.51&50.72&73.27\\
DWC w/o SSG&31.73&57.80&34.99&59.91&39.42&65.63&17.32&55.44&78.42\\
DWC w/o CLIP  &24.79&49.28&24.48&50.34&29.37&55.73&17.24&52.32&74.36\\
DWC&\textbf{32.67}&\textbf{57.96}&\textbf{35.53}&\textbf{60.11}&\textbf{40.13}&\textbf{66.09}&\textbf{18.94}&\textbf{55.55}&\textbf{80.19}\\
\hline
\end{tabular}}}
\vspace{-1mm}
\label{Stab1}
\end{table*}

\begin{table*}
\caption{Impact of pre-trained CLIP models with different backbones. }
\vspace{-2mm}
\centering
\setlength{\tabcolsep}{5mm}{
\resizebox{\linewidth}{!}{
\begin{tabular}{c|cc|cc|cc|ccc}
\hline
\multirow{2}{*}{CLIP}& \multicolumn{2}{c|}{Dress} & \multicolumn{2}{c|}{Shirt} & \multicolumn{2}{c|}{Toptee} & \multicolumn{3}{c}{Shoes} \\
&R@10&R@50 &R@10&R@50 &R@10&R@50 &R@1&R@10 &R@50\\
\hline
RN50&32.67&57.96&35.53&60.11&40.13&66.09&18.94&55.55&\textbf{80.19}\\
RN101&32.42&58.8&35.97&61.09&39.93&67.67&\textbf{19.31}&54.25&79.38\\
ViT-B32&31.63&57.85&34.99&60.50&38.55&66.4&19.00&\textbf{56.26}&79.64\\
ViT-B16&\textbf{33.61}&\textbf{58.80}&\textbf{37.09}&\textbf{62.46}&\textbf{40.80}&\textbf{68.38}&18.46&54.92&79.04\\
\hline
\end{tabular}}}
\vspace{-2mm}
\label{Stab3}
\end{table*}

\begin{table}[h]
\renewcommand\arraystretch{1.2}
\caption{Ablation study for different soft labels.}
\vspace{-2mm}
\centering
\setlength{\tabcolsep}{0.4mm}{
\resizebox{\linewidth}{!}{
\begin{tabular}{l|cc|cc|cc}
\hline
\multirow{2}{*}{Method}& \multicolumn{2}{c|}{Dress} & \multicolumn{2}{c|}{Shirt} & \multicolumn{2}{c}{Toptee} \\
&R@10&R@50 &R@10&R@50 &R@10&R@50 \\
\hline
Sigmoid  &31.82	&56.57	&33.81	&60.60	&38.70	&65.99\\
Euclidean distance &32.08	&57.66	&34.00	&60.75	&39.01	&65.63\\
Dot product (this paper) &\textbf{32.67}&\textbf{57.96}&\textbf{35.53}&\textbf{60.11}&\textbf{40.13}&\textbf{66.09}\\
\hline
\end{tabular}}}
\label{Rtab3}
\end{table}

\begin{table*}
\caption{Analysis of different combination methods to show the modality importance disparity and the necessary evidence of our model for improving the modality imbalance. }
\vspace{-2mm}
\centering
\setlength{\tabcolsep}{5mm}{
\resizebox{\linewidth}{!}{
\begin{tabular}{c|cc|cc|cc|ccc}
\hline
\multirow{2}{*}{Method}& \multicolumn{2}{c|}{Dress} & \multicolumn{2}{c|}{Shirt} & \multicolumn{2}{c|}{Toptee} & \multicolumn{3}{c}{Shoes} \\
&R@10&R@50 &R@10&R@50 &R@10&R@50 &R@1&R@10 &R@50\\
\hline
Image Only&7.83&20.43&13.74&26.64&10.20&24.22&8.24&31.92&57.31\\
Text Only&18.79&43.93&22.52&46.52&25.50&54.26&3.86&15.39&36.01\\
Summation&23.55&49.00&26.40&50.29&29.37&55.33&15.51&50.72&73.27\\
Concatenation&28.31&55.38&30.52&56.18&35.03&62.00&16.13&50.5&76.57\\
Naive Weight&28.36&54.44&30.30&56.62&34.98&62.16&15.17&50.16&75.66\\
our DWC&\textbf{32.67}&\textbf{57.96}&\textbf{35.53}&\textbf{60.11}&\textbf{40.13}&\textbf{66.09}&\textbf{18.94}&\textbf{55.55}&\textbf{80.19}\\
\hline
\end{tabular}}}
\label{Stab2}
\end{table*}

%
%
%
%

\subsection{2. Datasets}
To evaluate the practical value of our model, we chose three real-world datasets: Fashion200K \cite{Fashion200k2017}, Shoes \cite{shoes2018}, and FashionIQ \cite{wu2021fashioniq}. The details are described as follows.

\textbf{Fashion200k} \cite{Fashion200k2017} is a large-scale fashion dataset crawled from multiple online shopping websites, which includes more than 200K fashion clothing images of 5 different fashion categories, namely: \textit{pants, skirts, dresses, tops, jackets}. Each image has a human-annotated caption with accompanying attributes (e.g., ``black sleeveless printed ballgown''). Following ~\cite{vo2019composing}, we use the data split of about 172k images for training and  33,480 test queries for evaluation. Due to there is no already matched query image, query text, and target image, pairwise images with attribute-like modifications are generated by comparing their product descriptions. 
We structure the query text in the form of ``replace [sth] with [sth] and get [sth]'', e.g., ``replace tartan with floral and get red floral print shirt''.   \label{defination}

\textbf{Shoes} \cite{shoes2010} is a dataset originally crawled from the website of \textit{like.com}. It is further tagged with relative captions in natural language for dialog-based interactive retrieval \cite{shoes2018}, e.g., ``is shinier fire-engine red with higher platform''. Following \cite{shoes2018}, we use 10,000 training samples for training and 4,658 test samples for evaluation.

\textbf{FashionIQ} \cite{wu2021fashioniq} is a fashion retrieval dataset with interactive natural language captions, crawled from \emph{Amazon.com}. It consists of 77,684 images in total belonging to three categories: \textit{dresses}, \textit{top-tees} and \textit{shirts}. The dataset is organized by triplets, including a query image, a target image and a pair of relative captions that describe the differences between the two images, such as ``3/4 sleeve black and white dress and more top covered'' and ``has short sleeve and is black color''. Following the same evaluation protocol of VAL \cite{chen2020VAL}, we use the same training split and evaluate on the validation set. As a challenging dataset, only a training set of 18,000 triplets and a validation set of 6,016 triplets are available. We report the performance on the validation set as the test set labels are not available.

\subsection{3. Implementation Details}
We implement our method using PyTorch. Without loss of generality, for the CNN stream, we use the same backbone following most previous methods. Concretely, we adopt ResNet-50 as the image encoder and LSTM as the text encoder. For the CLIP stream, we take the publicly available pre-trained CLIP (RN50) model with an input size of $224 \times 224$ as the image and text encoder. The pre-trained CLIP is accessed at \url{https://github.com/openai/CLIP} without fine-tuning. We use an SGD optimizer with a learning rate set to $1e-4$ and train the model for a maximum of 50 epochs. We empirically set the trade-off parameter in Eq. 14 in the main body as $\lambda=0.8$. The batch size $B=32$.
We use retrieval accuracy $R@N$ as our evaluation metric, by computing the percentage of test queries where at least one target or correctly searched image is within the top $N$ retrieved images. We also present the mean precision when $N$ is set as different values.

\subsection{4. Ablation Study and Model Analysis}
\textbf{Different formulas to generate soft labels.} We explore and experiment with different formulas to generate soft labels for the training data. The experimental results are shown in Table \ref{Rtab3}. \emph{Sigmoid} means replacing the Softmax function in Eq. (12) with Sigmoid function to generate soft labels, whose value ranges from 0 to 1. \emph{Euclidean distance} and \emph{Dot product} take the same form of Eq. (12), where $\kappa (\cdot,\cdot)$ takes Euclidean distance or dot product as the similarity measure formula, respectively. For brevity, we choose to present the most effective formula in the paper.

\textbf{Ablation Study and Model Analysis with More Evaluation Metrics}. Due to the space limitation, we only present the experimental results of R@10 in the ablation study and model analysis (\textbf{Section 4.2} of the main body). We further enrich Tables 5, 6 and 7 of the main body by providing more evaluation results of R@1 and R@50 metrics. The contributions of different components in our proposed model are presented in Table \ref{Stab1} (Supplementary data for Table 5 in the main body). The impact of pre-trained CLIP models under different backbones is provided in Table \ref{Stab3} (Supplementary data for Table 7 in the main body).
The modality importance disparity by testing different combination methods to show the necessity of our EMD is presented in Table \ref{Stab2} (Supplementary data for Table 6 in the main body).

\subsection{5. Retrieval performance under different number of retrieved samples}
\begin{figure}[h]
\centering
\vspace{-2mm}
 \centerline{\includegraphics[width=0.4\textwidth]{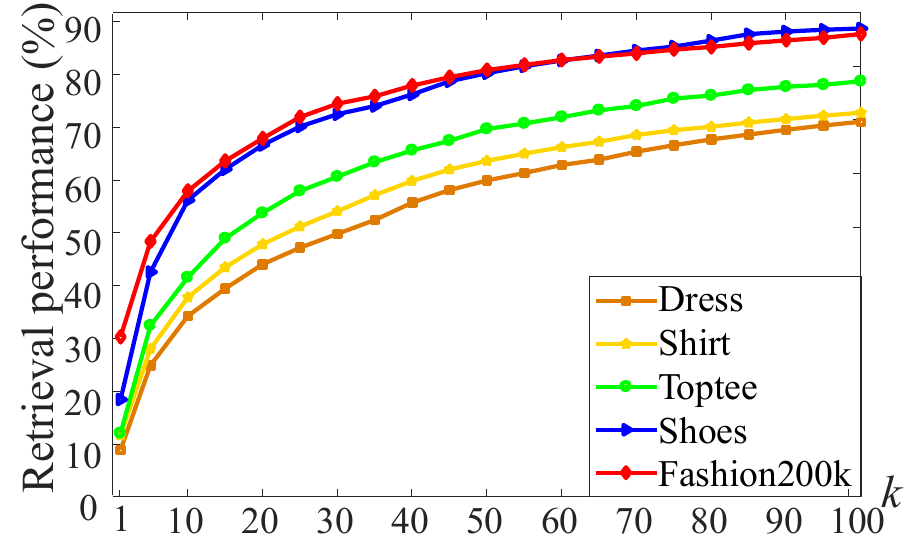}}
\caption{Retrieval performance (\%) with the number of retrieved samples (k) varying from 1 to 100.}
\vspace{-2mm}
\label{topk}
\end{figure}

We have provided the retrieval performance with different $k$ value varying from 1 to 100  in Fig. \ref{topk}. We can observe that as the value of $k$ increases, the retrieval performance gets better because the probability of retrieving the correct sample is increased.

\subsection{6. More Visualization Results with Rationality and Interpretability of Our Approach}
\begin{figure*}
\centering
\includegraphics[width=0.9\textwidth]{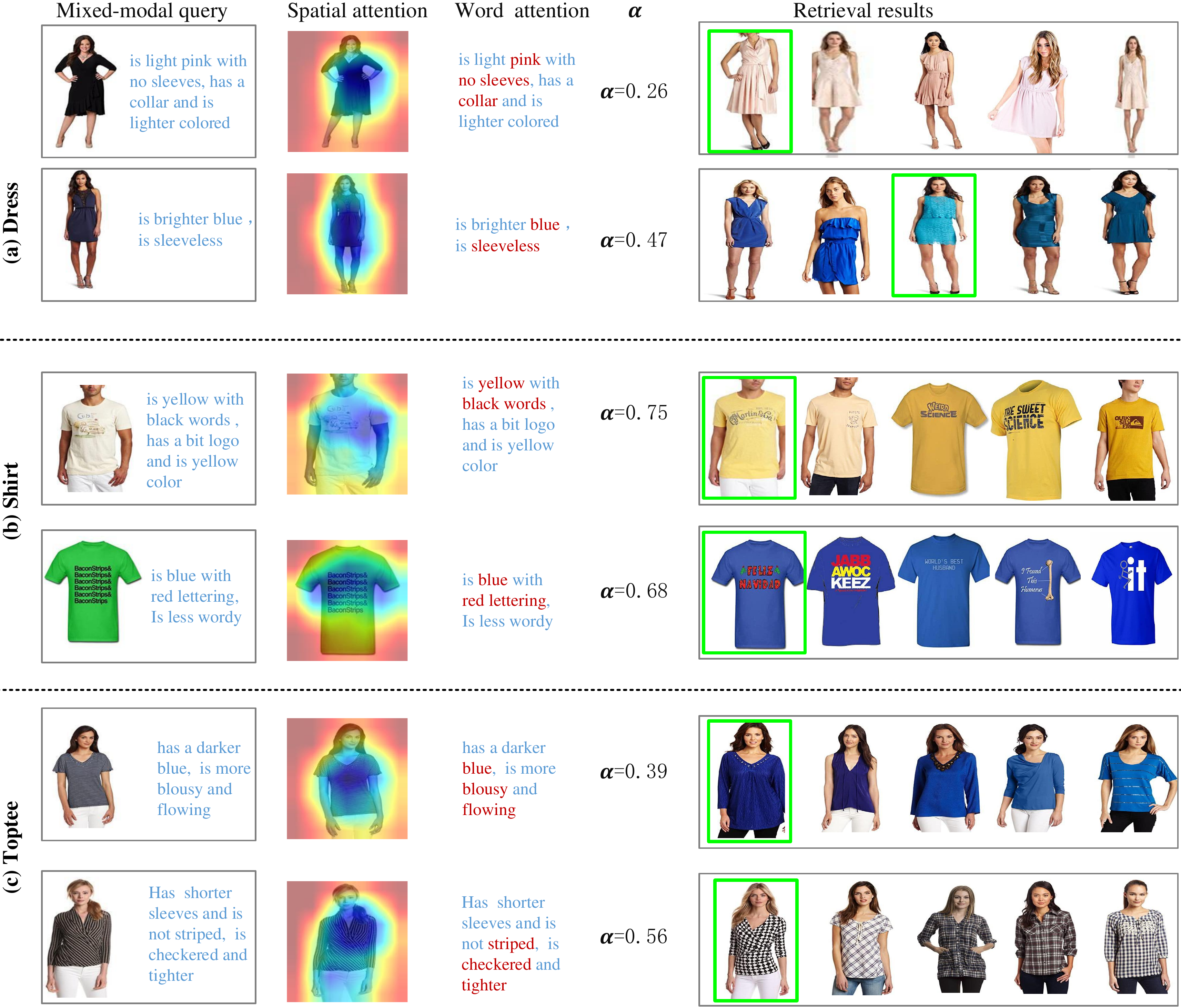}
\caption{Visualization results with more rationality and interpretability, which contain attention visualization, modality importance disparity and retrieval results. $\alpha$ is the importance of image modality in Eq.11 of the main body. Ground-truths target images are shown in green boxes.}
\label{Sfig3}
\vspace{-1.2mm}
\end{figure*}

To further investigate the properties of the proposed model, we present qualitative results in Fig.~\ref{Sfig3}, which contains attention visualization, learned modality importance and the Top-5 retrieval results. From the results, the effectiveness of the proposed model is well interpreted. Take Fig.~\ref{Sfig3}(a) as an example, the first row means that the text modality is more important than the image modality, and the learned image modality importance weight is $\alpha=0.26$ (In Eq.11 of the main body), which is smaller than $1-\alpha=0.74$ of text modality. The second row means that both modalities have similar importance, and therefore the learned image modality importance weight is $\alpha=0.47$, which is comparable to $1-\alpha=0.53$ of text modality.
For other examples, we can also observe the rationality and interpretability of the learned importance weights, heatmaps and retrieval results, according to the given image and text descriptions.

\end{document}